\documentclass[conference]{IEEEtran}
\IEEEoverridecommandlockouts
\usepackage{cite}
\usepackage{amsmath,amssymb,amsfonts}
\usepackage{algorithm}
\usepackage{algpseudocode}
\usepackage{graphicx}
\usepackage{url}
\usepackage{hyperref}
\usepackage{textcomp}
\usepackage{booktabs}
\usepackage{xcolor}
\DeclareMathAlphabet{\mathpzc}{OT1}{pzc}{m}{it}
\usepackage{placeins}
\usepackage{multirow}
\usepackage{subcaption}
\usepackage{breqn}
\usepackage{enumitem}
\usepackage{array}
\usepackage{tabularx}
\usepackage{amsthm}
\def\BibTeX{{\rm B\kern-.05em{\sc i\kern-.025em b}\kern-.08em
    T\kern-.1667em\lower.7ex\hbox{E}\kern-.125emX}}
\usepackage{color, colortbl}	
\definecolor{LightCyan}{rgb}{0.88,1,1}
\definecolor{Gray}{gray}{0.90}

\begin{document}

\title{Initial Exploration of Zero-Shot Privacy Utility Tradeoffs in Tabular Data Using GPT-4}
\author{\IEEEauthorblockN{\textsuperscript{}Bishwas Mandal}
\IEEEauthorblockA{\textit{Department of Computer Science} \\
\textit{Kansas State University}\\
 Manhattan, KS 66506, USA \\
 bishmdl76@ksu.edu}
\and
\IEEEauthorblockN{\textsuperscript{}George Amariucai}
 \IEEEauthorblockA{\textit{Department of Computer Science} \\
\textit{Kansas State University}\\
 Manhattan, KS 66506, USA \\
 amariucai@ksu.edu}
\and
\IEEEauthorblockN{\textsuperscript{}Shuangqing Wei}
\IEEEauthorblockA{\textit{Division of Electrical \& Computer Engineering} \\
\textit{Louisiana State University}\\
 Baton Rouge, LA 70803, USA \\
 swei@lsu.edu}
}

\maketitle
\begin{abstract}

We investigate the application of large language models (LLMs), specifically GPT-4, to scenarios involving the tradeoff between privacy and utility in tabular data. Our approach entails prompting GPT-4 by transforming tabular data points into textual format, followed by the inclusion of precise sanitization instructions in a zero-shot manner. The primary objective is to sanitize the tabular data in such a way that it hinders existing machine learning models from accurately inferring private features while allowing  models to accurately infer utility-related attributes. We explore various sanitization instructions. Notably, we discover that this relatively simple approach yields performance comparable to more complex adversarial optimization methods used for managing privacy-utility tradeoffs. Furthermore, while the prompts successfully obscure private features from the detection capabilities of existing machine learning models, we observe that this obscuration alone does not necessarily meet a range of fairness metrics.  Nevertheless, our research indicates the potential effectiveness of LLMs in adhering to these fairness metrics, with some of our experimental results aligning with those achieved by well-established adversarial optimization techniques.
\end{abstract}


\section{Introduction}
The rapid growth observed in the deep learning field can be attributed to a range of factors, such as advancements in architectural designs, improved convergence capabilities, and the widespread availability of open-source models, among others. However, the primary catalyst driving the significant growth of deep learning is the accessibility of extensive datasets and the continuous enhancement of computational capabilities \cite{Sarker2021DeepLA}. One specific architectural advancement has made a substantial contribution to the effective utilization of GPUs for efficient computational usage: the \textit{Transformer} architecture \cite{NIPS2017_3f5ee243}. The transformer architecture serves as the foundational framework for numerous recent advancements, including the development of various Large Language Models (LLMs). These LLMs have popularized the practice of fine-tuning language models to a much greater extent than in the past.

LLMs are characterized by their massive architecture, with models having billions of parameter (e.g GPT-3 model with 175 billion parameters \cite{brown2020language}). Training these models involves working with extensive datasets that tap into a wide array of information from diverse data sources. By capitalizing on insights derived from these varied sources, LLMs have a tendency to amass a comprehensive range of knowledge, rendering them highly adaptable to a diverse array of tasks. Therefore, they have found application across various domains, including applications in sentiment analysis, text generation, code generation, multimodal tasks, and more. They have also proven effective in wide array of tasks in low-data scenarios such as zero-shot and few-shot settings \cite{brown2020language}. LLMs have not only demonstrated their effectiveness in handling unstructured data but have also exhibited remarkable performance when applied to structural tabular datasets within zero-shot and few-shot settings \cite{hegselmann2023tabllm}. 


LLMs pose a significant challenge when it comes to preserving privacy. One of the primary concerns is their inadvertent potential to reveal information about the training data  \cite{carlini2021extracting}. This vulnerability implies that malicious individuals could potentially access the training data, which may contain sensitive information, leading to privacy breaches. Additionally, LLMs have made substantial advancements in their ability to draw inferences, potentially allowing them to deduce various personal attributes of users.  \cite{staab2023memorization} demonstrate how large language models can deduce sensitive information from users during the inference phase. Therefore, privacy issues regarding LLMs go beyond just extracting training data; they also involve privacy violations due to LLMs' powerful inference capabilities.

Several studies \cite{10031034, 10.1145/3539597.3575792} have proposed methods,  to protect LLMs from revealing their training data. These methods focus on modifying the model's parameters using different techniques to enhance privacy. However, as of our current knowledge, there hasn't been prior research exploring whether LLMs can effectively be used to safeguard user privacy concerning their ability to make inferences. Therefore, our paper shifts its focus towards investigating the potential of LLMs to leverage their inference abilities and statistical insights to maintain user privacy, especially when dealing with tabular datasets. Our main objective is to use LLMs, especially GPT-4, to sanitize datasets in a way that hinders the extraction of sensitive user information while still retaining the ability to extract useful features. To the best of our knowledge, this study represents the first attempt to explore how LLMs can be employed to enhance the privacy of tabular datasets while preserving their usefulness.

\begin{figure*}[!htb]
    \centering
    \includegraphics[width=0.99\textwidth]{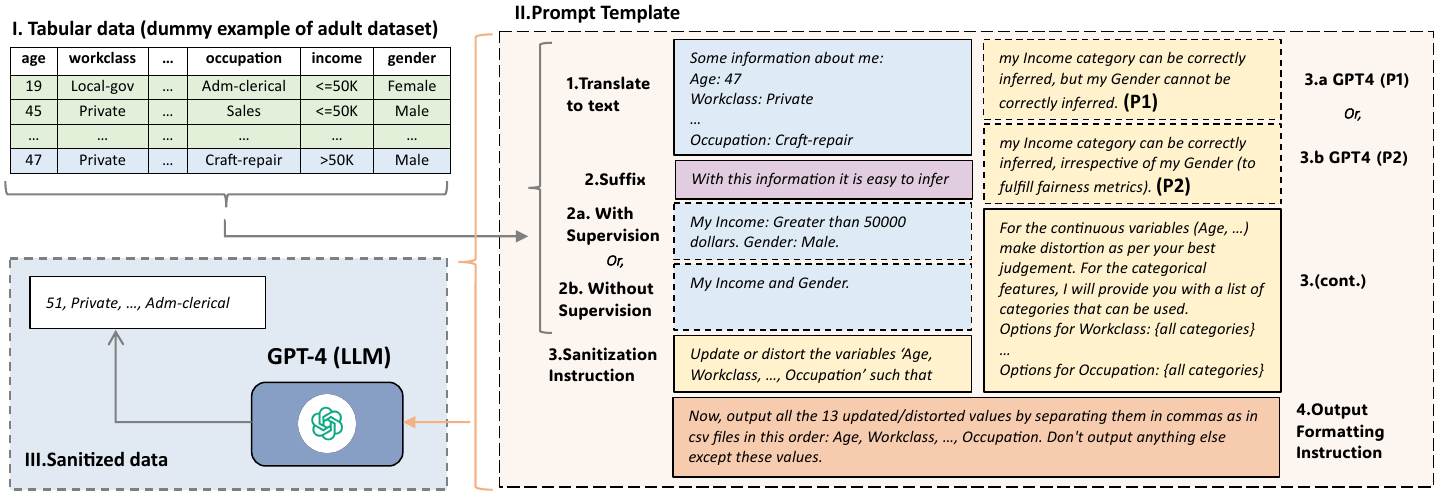}
    \caption{\textmd{Overview of the proposed data sanitization technique using GPT-4.  Our approach involves initializing the GPT-4 model with a prompt that includes the conversion of tabular data points into textual format, followed by the incorporation of sanitization instructions. Additionally, we incorporate formatting instructions into the prompt to ensure that the generated outputs are suitable for programmatic utilization. The presented prompt above pertains to the prompt designated for \textit{Task 1}.}}
    \vspace{-0.1in}
    \label{fig:gpt4-privacy overview}
\end{figure*} 

\section{Related Work}
Within the sphere of privacy, two fundamental categories have been established: \textit{data privacy} and \textit{inference privacy} \cite{8259344, 8766142}.  \textit{Data privacy} primarily focuses on preserving the original, unaltered data in its raw state. \textit{Inference privacy} deals with the protection of sensitive information that can be derived or deduced from the disclosed raw data, often as a consequence of correlations. In this context, \textit{raw data} denotes the initial data collected or generated, which remains unaltered and represents the data in its fundamental state before undergoing any form of examination, modification, or inference.

In the broader discourse concerning privacy implications in LLMs, a central concern revolves around the identification of the training data. This aligns with the paradigm of \textit{data privacy}, wherein the actual raw data contains sensitive information, and the extraction of this data poses a substantial privacy risk. \cite{carlini2021extracting, nasr2023scalable, li2023mope}, show diverse methods for extracting training data from language models. To mitigate the risk of extraction of private training data, \cite{10031034} introduce a method for privately fine-tuning large language models using Differential Privacy, and \cite{10.1145/3539597.3575792} further demonstrate the fine-tuning of LLMs with enhanced privacy. In a similar vein, \cite{wu2023depn} propose a privacy neuron detector designed to identify neurons associated with private information and subsequently modify these identified privacy neurons by setting their activations to zero. Furthermore, \cite{kassem-etal-2023-preserving} enhance the privacy of language models through reinforcement learning, utilizing negative similarity scores to promote paraphrasing and reduce dependence on the original training data. \cite{huang2023privacy} utilize a straightforward sanitization step aimed at mitigating privacy risks in retrieval-based language models.

It is important to note that these methods primarily address the issue of \textit{data privacy} and do not directly tackle the core focus of our research, which centers on \textit{inference privacy}. \cite{staab2023memorization} demonstrate privacy concerns related to LLMs extend beyond mere memorization of training data and how the improved efficiency of LLMs in inference can be leveraged to infer private attributes from user texts during inference. While LLMs have predominantly focused on privacy from an unstructured data perspective (e.g text, image, etc.), \cite{hegselmann2023tabllm} show LLMs can be utilized for classification tasks in zero-shot and few-shot settings in tabular datasets as well. The statistical knowledge acquired by LLMs extends beyond textual data, potentially posing privacy risks in tabular datasets, as demonstrated by \cite{staab2023memorization}, where GPT-4 is capable to infer private attributes such as birth place, race, education level, income, and gender with a high accuracy from the ACS Income Dataset \cite{ding2022retiring} in a zero-shot setting.

Various adversarial optimization techniques \cite{edwards2016censoring, e19120656, madras2018learning, pittaluga2018learning, 8515092, huang2019generative, chen2019distributed, wu2020privacypreserving, morales2020sensitivenets, Xiao_Tsai_Sohn_Chandraker_Yang_2020, erdemir2021active, wu2021privacypreserving, 9892789}, have been utilized to tackle the inherent trade-offs between safeguarding inference privacy and preserving data utility. These techniques either introduce additional noise into their generator network or latent variables or employ loss functions that achieve distortion without explicitly adding noise. They have proven their effectiveness in thwarting Machine Learning (ML) models from inferring sensitive attributes like race, gender, income, and others depending on the application. Moreover, these methods are adaptable and can be applied to different data types, including tables, images, and text (representations). The primary focus of these methodologies has revolved around ensuring inference privacy, with the aim of preventing the revelation of sensitive attributes or meeting fairness criteria within ML models. However, as pointed out by \cite{9892789}, a notable portion of this research has not thoroughly addressed the unique characteristics of the data and the requisite post-processing steps when dealing with tabular datasets that incorporate categorical features. They emphasize that sanitized datasets generated through diverse privacy mechanisms often necessitate additional post-processing efforts to restore the data to its original format.


\section{Problem Formulation and Methodology}

\textbf{Problem:} Let's consider the scenario of an external service provider tasked with the responsibility of sanitizing a dataset to both protect user privacy and ensure the desired utility is preserved. This provider has access to a raw database containing a data vector, denoted as $D$, which consists of various features represented as $\{d_1, d_2, d_3, \ldots, d_n\}$. The primary objective for this provider is to sanitize the dataset $D$ in such a way that when users release their sanitized data, denoted as $\hat{D}$, it enables accurate inference of utility features, denoted as $D_U$, while simultaneously making it difficult to accurately infer private features, denoted as $D_P$. Initially, the dataset $D$ exhibits correlations with both $D_U$ and $D_P$ and it's important to note that $D_U \notin D$ and $D_P \notin D$. Failing to properly sanitize and releasing $D$ publicly could lead to the accurate inference of private features, thus compromising user privacy. In a public domain, potential attackers might utilize various pre-trained machine learning models to deduce private features from the data. Therefore, it is crucial for the sanitizing function $f$ to be robust against multiple models to ensure effective privacy protection. In the context of this paper, \emph{privacy is construed as the pretrained model's incapacity to accurately deduce private attributes, whereas utility is characterized as the pretrained model's competence in accurately deducing utility attributes}.

Prior research has primarily focused on adversarial optimization approaches to identify an effective sanitizing function $f$, or other theoretical techniques involving noise addition \cite{9500410}. In contrast, our approach explores the potential of using GPT-4 as a viable sanitizing function $f$.

\textbf{Proposed Methodology:} We now detail the methodology for employing GPT-4 as a sanitizing function $f$. In this specific context, the function $f$ is a composite of two distinct functions, namely $p$ and $g$, expressed as $f(.) = g(p(.))$. Here, $p$ is responsible for the prompting mechanism, while $g$ represents the pre-trained GPT-4 model. The prompting function $p$ operates by taking the original tabular data point as its input and generating a prompt. The prompting function $p$ encompasses several sequential steps elaborated below:

\begin{enumerate}
    \item \textbf{\textit{Translation to Text}}: The first phase involves converting tabular data into text. \cite{hegselmann2023tabllm} demonstrate that different techniques for this conversion produce similar outcomes, as long as all the information from the tabular data is retained. Therefore, we adopt a straightforward technique for this conversion. The exact prompt is illustrated in Fig. \ref{fig:gpt4-privacy overview}, where the variables \textit{\{age, workclass, ..., occupation\}} represent all features in vector $D$.

    \item \textbf{\textit{Suffix with Supervision}}: This stage involves annotating the data $D$ with accurate private and utility labels $D_P$ and $D_U$, respectively. This annotated information is subsequently incorporated into the prompt, as depicted in Fig. \ref{fig:gpt4-privacy overview}. In this illustration, $D_P$ corresponds to gender, while $D_U$ corresponds to income.

    \item \textbf{\textit{Sanitization Instruction}}: This portion involves the addition of prompts that instruct the model to sanitize the dataset $D$ in order to achieve the specified objectives related to privacy and utility. We conducted an evaluation using two distinct prompts, namely $P1$ and $P2$. The precise prompts are displayed in Fig. \ref{fig:gpt4-privacy overview}.

    \item \textbf{\textit{Output Formatting Instruction}}: The final part of the prompt entails incorporating instructions for arranging the output in a specific sequence, ensuring that the outcome from the GPT-4 function $g$ aligns with our desired format and can be seamlessly added to the database as $\hat{D}$.
\end{enumerate}

Upon completion of the function $p$ transforming the tabular data into a comprehensive prompt that encompasses all necessary guidelines for data sanitization, the resulting output of $p$ is then supplied to the function $g$. The role of function $g$ is to produce the sanitized dataset, denoted as $\hat{D}$. In formal terms, this process is represented by the equation $\hat{D} = g(p(D, D_U, D_P))$, indicating the sequential application of functions $p$ and $g$ to achieve the final sanitized dataset.

\subsection{Assumptions and Threat Model}

In our conceptual framework, we establish a foundational trust relationship between the data owner and the third-party service provider. This trust empowers the data owner to confidently share raw data with the service provider, who is responsible for data sanitization. Analysts or potential attackers seeking to extract private or utility features from the data $\hat{D}$ possess an auxiliary dataset at their disposal. This auxiliary dataset enables them to initially train various machine learning models that can predict the private and utility features. After these models are trained, analysts proceed to attempt the inference of private and utility features for the users whose data is contained within $\hat{D}$.

\subsection{Existing Sanitization Mechanisms (Baselines)}\label{different-models}
\begin{figure}[!htb]
\vspace{-0.2in}
    \centering
    \includegraphics[width=0.45\textwidth]{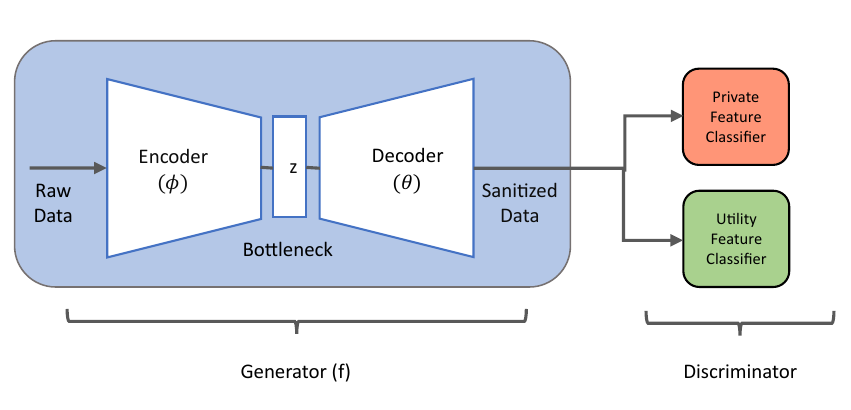}
    \caption{\textmd{ALFR and UAE-PUPET architecture}}
    \label{adversarial techniques}
    \vspace{-0.05in}
\end{figure} 

In order to evaluate the effectiveness of GPT-4 as a privacy-preserving tool, we draw comparisons with two established privacy mechanisms that employ adversarial optimization techniques: ALFR \cite{edwards2016censoring}, and UAE-PUPET \cite{9892789}. These methods are characterized by a generator-discriminator framework, depicted in Fig. \ref{adversarial techniques}. Within this architecture, the generator is a generative model comprising an encoder-decoder pair, while the discriminator is made up of neural network models tasked with classifying private and utility features. The discriminator's role is crucial, providing the necessary loss function that ensures the sanitized data $\hat{D}$ conceals private features while maintaining the inferability of utility features.

For our analysis, we employ the official implementation \cite{githubGitHubBishwasmandal246uaepupet} of UAE-PUPET to produce the results presented in our paper. In the case of ALFR, we adopt the sanitization algorithm outlined in their paper and integrate it with the model architecture used in UAE-PUPET to ensure consistency.  Algorithm \ref{alg:existing} demonstrates the ALFR sanitization mechanism. In this algorithm, $f$, $c_p$, and $c_u$ correspond to functions associated with generator, private feature classifier, utility feature classifier networks. In parallel $\gamma$, $\gamma_p$, and $\gamma_u$ denote the weights of these functions, respectively. Additionally, $\alpha$, $\lambda_p$, and $\lambda_u$ are scalars utilized to prioritize specific loss components in the final loss function L.

\begin{algorithm}[!htb]
\small
\caption{ALFR training algorithm}\label{alg:existing}
\begin{algorithmic}
\State $\gamma, \gamma_p, \gamma_u \gets$  initialize weights
\State $U \gets True$ 
\Repeat
\State $\hat{D} \gets f(D)$, $\hat{D}_P \gets c_p(\hat{D})$, $\hat{D}_U \gets c_u(\hat{D})$
\State $C \gets \text{MSE}(\hat{D}, D)$ 

\State $l_p \gets \text{cross\_entropy}(\hat{D}_P, D_P)$ 

\State $l_u \gets \text{cross\_entropy}(\hat{D}_U, D_U)$ 

\State $L \gets \alpha C - \lambda_{p} l_p + \lambda_{u} l_u$ 

\If {$U$}
\State Update $\gamma_p$ to \emph{maximize} the loss L
\Else
\State Update ($\gamma, \gamma_u$) to \emph{minimize} the loss L
\EndIf
\State $U \gets$ not $U$
\Until Deadline
\end{algorithmic}
\end{algorithm}
\vspace{-0.135in}

\section{Experimental Setup}

\subsection{Dataset}
We conduct our experiment using the UCI Adult dataset \cite{misc_adult_2}. Our dataset selection process involved reviewing various tabular datasets initially used for fairness studies, such as \textit{COMPAS, Bank Marketing, Communities and Crime, Student Performance, and German Credit Dataset}, as well as others mentioned in \cite{hegselmann2023tabllm} like \textit{Blood, Callhousing, Car, Diabetes, Heart, and Jungle}. The selection of the UCI Adult dataset was motivated by several considerations. Firstly, we required a dataset that encompassed a minimum of two dependent variables capable of serving as private and utility features. Secondly, our research objective involved the development of machine learning models that would not be able to accurately infer private features following data sanitization. Consequently, it was imperative to have at least two dependent variables with high initial accuracy to enable us to assess whether there was a notable alteration in model accuracy subsequent to data sanitization.

Furthermore, we sought a dataset with high zero-shot classification scores when utilized with LLMs, signifying that the language model possess statistical information pertaining to that dataset. To fulfill all these criteria, the UCI Adult dataset emerged as the only suitable choice. Although the ACS Income Dataset met these criteria as well, we ultimately opted exclusively for the UCI Adult dataset due to its similar nature and the fact that the inclusion of the ACS dataset would not significantly diversify our experiments. In order to introduce diversity into our research endeavors, we devised two distinct tasks, designated as Task 1 and Task 2.

\subsubsection*{\textbf{Task 1}}
Task 1 focuses on the UCI Adult dataset, designating Gender as the private feature and the prediction of whether an individual's income exceeds \$50,000 as the utility feature, similar to the work in \cite{9892789}. In this task, all features apart from the private and utility ones are classified as variables requiring sanitization, denoted as $D$ in our problem formulation.

\subsubsection*{\textbf{Task 2}}

In Task 2, although the features designated for sanitization ($D$) stay the same, there's a role reversal between the private and utility features. In this context, the private feature is now the classification task determining whether an individual's income exceeds \$50,000, and the Gender is considered the utility feature. This modification is intentionally made to demonstrate that the outcomes of our research are not influenced by the specific selection of a feature as private or another as utility and therefore, confirms the effectiveness of our proposed methodology, irrespective of which variable is designated as private or utility.

\subsection{Performance Evaluation Metrics}

To evaluate the efficacy of our sanitization mechanism, we proceed to compute the accuracy and F1 scores for an array of classifier models. Initially, we calculate these scores using the dataset in its unsanitized state. Subsequently, we assess the performance of these models after the data has undergone sanitization. Our expectation is that the accuracy and F1 scores for the utility feature should exhibit minimal decline in comparison to their original scores when the data remains unsanitized. However, for the private feature, our objective is to achieve a significant reduction in these scores.

Additionally, we employ privacy-utility tradeoff metrics introduced in \cite{10030960}. They define privacy leakage for attribute $p$ as follows:
\begin{equation} \label{eq1}
M_p = \frac{c_a(p) - c_r(p)}{c_n(p) - c_r(p)}
\end{equation}

Similarly, utility performance of attribute $u$, is defined as follows:
\begin{equation} \label{eq2}
M_u = \frac{c_a(u) - c_r(u)}{c_n(u) - c_r(u)}
\end{equation}

We ensure that these metrics are constrained within the range of 0 to 1. In Eq. \ref{eq1}, Eq. \ref{eq2}, $c_n(.)$ represents the accuracy of a classifier in predicting either the private feature $p$ or the utility feature $u$ using raw, unsanitized data. $c_a(.)$ denotes the accuracy after the sanitization process, while $c_r(.)$ signifies the accuracy associated with random guessing. A lower value of $M_p$ indicates enhanced privacy, as it implies a closer resemblance to random guessing. Conversely, a higher value of $M_u$ is preferable, signifying a lower drop in utility. It is noteworthy that the accuracy for random guessing in predicting Income is 0.74, while for Gender, it is 0.69 (these are unbalanced classification problems).

\begin{table*}[!htb]
\centering
\begin{tabular}{m{1.6cm}m{1.4cm}cccccccc}
\toprule
\multirow{4}{*}{\textbf{Privacy}} & \multirow{4}{*}{\textbf{Models}} & \multicolumn{4}{c}{\textbf{Task 1}} & \multicolumn{4}{c}{\textbf{Task 2}} \\
\cmidrule(lr){3-6} \cmidrule(lr){7-10}
& & \multicolumn{2}{c}{\textbf{Gender (Private)}} & \multicolumn{2}{c}{\textbf{Income (Utility)}} & \multicolumn{2}{c}{\textbf{Income (Private)}} & \multicolumn{2}{c}{\textbf{Gender (Utility)}} \\
\cmidrule(lr){3-4} \cmidrule(lr){5-6} \cmidrule(lr){7-8} \cmidrule(lr){9-10}
& & \textbf{Acc} & \textbf{F1} & \textbf{Acc} & \textbf{F1} & \textbf{Acc} & \textbf{F1} & \textbf{Acc} & \textbf{F1} \\
\midrule
\multirow{6}{=}{No PM} & LR & $0.83_{.002}$ & $0.84_{.002}$ & $0.85_{.003}$ & $0.84_{.003}$ & $0.85_{.003}$ & $0.84_{.003}$ & $0.83_{.002}$ & $0.84_{.002}$ \\
& RF & $0.84_{.002}$ & $0.84_{.003}$ & $0.87_{.002}$ & $0.85_{.003}$ & $0.87_{.002}$ & $0.85_{.003}$ & $0.84_{.002}$ & $0.84_{.003}$ \\
& XGB & $0.84_{.002}$ & $0.84_{.003}$ & $0.88_{.003}$ & $0.87_{.003}$ & $0.88_{.003}$ & $0.87_{.003}$ & $0.84_{.002}$ & $0.84_{.003}$ \\
& NN & $0.83_{.002}$ & $0.81_{.002}$ & $0.85_{.003}$ & $0.84_{.003}$ & $0.85_{.003}$ & $0.84_{.003}$ & $0.83_{.002}$ & $0.81_{.002}$ \\
& GPT-4 (C) & $0.80_{.002}$ & $0.78_{.002}$ & $0.80_{.003}$ & $0.80_{.003}$ & $0.80_{.003}$ & $0.80_{.003}$ & $0.80_{.002}$ & $0.78_{.002}$ \\
& \textit{Summary} & $0.84_{.002}$ & $0.84_{.002}$ & $0.88_{.003}$ & $0.87_{.003}$ & $0.88_{.003}$ & $0.87_{.003}$ & $0.84_{.002}$ & $0.84_{.002}$ \\
\midrule
\multirow{6}{=}{ALFR} & LR & $0.65_{.006}$ & $0.64_{.007}$ & $0.81_{.006}$ & $0.79_{.004}$ & $0.75_{.007}$ & $0.70_{.005}$ & $0.81_{.003}$ & $0.81_{.009}$ \\
& RF & $0.65_{.010}$ & $0.63_{.009}$ & $0.81_{.007}$ & $0.79_{.006}$ & $0.75_{.006}$ & $0.68_{.008}$ & $0.80_{.005}$ & $0.80_{.005}$ \\
& XGB & $0.64_{.005}$ & $0.63_{.007}$ & $0.81_{.008}$ & $0.80_{.005}$ & $0.72_{.007}$ & $0.68_{.009}$ & $0.80_{.006}$ & $0.80_{.007}$ \\
& NN & $0.65_{.007}$ & $0.65_{.008}$ & $0.81_{.010}$ & $0.79_{.008}$ & $0.75_{.006}$ & $0.71_{.007}$ & $0.79_{.004}$ & $0.80_{.007}$ \\
& GPT-4 (C) & $0.65_{.006}$ & $0.62_{.007}$ & $0.75_{.005}$ & $0.75_{.005}$ & $0.66_{.008}$ & $0.67_{.012}$ & $0.78_{.007}$ & $0.75_{.011}$ \\
&  \textit{Summary} & $\mathbf{0.65}_{.006}$ & $\mathbf{0.65}_{.008}$ & $0.81_{.006}$ & $0.80_{.005}$ & $0.75_{.006}$ & $0.71_{.007}$ & $0.81_{.003}$ & $0.81_{.009}$ \\
\midrule
\multirow{6}{=}{UAE-PUPET} & LR & $0.67_{.009}$ & $0.65_{.008}$ & $0.79_{.007}$ & $0.78_{.005}$ & $0.73_{.011}$ & $0.67_{.005}$ & $0.81_{.005}$ & $0.81_{.004}$ \\
& RF       & $0.67_{.006}$ & $0.64_{.011}$ & $0.80_{.006}$ & $0.79_{.006}$ & $0.74_{.008}$ & $0.67_{.006}$ & $0.81_{.004}$ & $0.81_{.006}$ \\
& XGB             & $0.66_{.008}$ & $0.64_{.010}$ & $0.78_{.004}$ & $0.77_{.004}$ & $0.72_{.006}$ & $0.69_{.009}$ & $0.82_{.005}$ & $0.82_{.003}$ \\
& NN                  & $0.66_{.007}$ & $0.64_{.008}$ & $0.79_{.008}$ & $0.78_{.008}$ & $0.71_{.006}$ & $0.68_{.009}$ & $0.81_{.007}$ & $0.81_{.008}$ \\
& GPT-4 (C)        & $0.65_{.008}$ & $0.60_{.007}$ & $0.75_{.006}$ & $0.75_{.006}$ & $0.60_{.009}$ & $0.62_{.010}$ & $0.78_{.006}$ & $0.75_{.006}$ \\
& \textit{Summary} & $0.67_{.006}$ & $\mathbf{0.65}_{.008}$ & $0.80_{.006}$ & $0.79_{.006}$ & $0.74_{.008}$ & $0.69_{.009}$ & $\mathbf{0.82}_{.005}$ & $\mathbf{0.82}_{.003}$ \\
\midrule
\multirow{6}{=}{GPT4 (P1)} & LR & $0.36_{.006}$ & $0.30_{.005}$ & $0.88_{.004}$ & $0.88_{.004}$ & $0.46_{.008}$ & $0.49_{.008}$ & $0.81_{.004}$ & $0.82_{.007}$ \\
& RF       & $0.65_{.005}$ & $0.66_{.007}$ & $0.79_{.006}$ & $0.74_{.005}$ & $0.63_{.005}$ & $0.58_{.010}$ & $0.81_{.008}$ & $0.81_{.006}$ \\
& XGB          & $0.35_{.008}$ & $0.55_{.006}$ & $0.86_{.005}$ & $0.86_{.005}$ & $0.67_{.006}$ & $0.67_{.006}$ & $0.79_{.007}$ & $0.80_{.007}$ \\
& NN                  & $0.47_{.008}$ & $0.46_{.005}$ & $0.79_{.005}$ & $0.80_{.005}$ & $0.47_{.011}$ & $0.50_{.009}$ & $0.79_{.004}$ & $0.80_{.005}$ \\
& GPT-4 (C)       & $0.60_{.007}$ & $0.61_{.005}$ & $0.89_{.004}$ & $0.89_{.005}$ & $0.35_{.013}$ & $0.39_{.009}$ & $0.79_{.006}$ & $0.78_{.007}$ \\
& \textit{Summary} & $\mathbf{0.65}_{.005}$ & $0.66_{.007}$ & $\mathbf{0.89}_{.004}$ & $\mathbf{0.89}_{.005}$ & $\mathbf{0.67}_{.006}$ & $\mathbf{0.67}_{.006}$ & $0.81_{.004}$ & $\mathbf{0.82}_{.007}$ \\
\midrule
\multirow{6}{=}{GPT4 (P2)} & LR & $0.75_{.004}$ & $0.76_{.005}$ & $0.88_{.005}$ & $0.88_{.005}$ & $0.71_{.006}$ & $0.72_{.008}$ & $0.79_{.003}$ & $0.80_{.005}$ \\
& RF       & $0.73_{.004}$ & $0.74_{.007}$ & $0.87_{.007}$ & $0.87_{.006}$ & $0.72_{.011}$ & $0.73_{.009}$ & $0.78_{.006}$ & $0.79_{.007}$ \\
& XGB           & $0.74_{.008}$ & $0.75_{.007}$ & $0.85_{.005}$ & $0.86_{.005}$ & $0.74_{.011}$ & $0.75_{.012}$ & $0.77_{.007}$ & $0.78_{.007}$ \\
& NN                  & $0.73_{.006}$ & $0.74_{.006}$ & $0.86_{.008}$ & $0.86_{.009}$ & $0.69_{.010}$ & $0.71_{.007}$ & $0.78_{.005}$ & $0.78_{.005}$ \\
& GPT-4 (C)       & $0.73_{.004}$ & $0.73_{.005}$ & $0.83_{.005}$ & $0.83_{.006}$ & $0.62_{.008}$ & $0.64_{.010}$ & $0.77_{.005}$ & $0.77_{.006}$ \\
& \textit{Summary} & $0.75_{.004}$ & $0.76_{.005}$ & $0.88_{.005}$ & $0.88_{.005}$ & $0.74_{.011}$ & $0.75_{.012}$ & $0.79_{.003}$ & $0.80_{.005}$ \\
\bottomrule
\end{tabular}
\caption{\textmd{
Accuracy and F1 scores of multiple machine learning models evaluated both before and after the application of the sanitization mechanism. No PM denotes the scores obtained prior to the implementation of any sanitization mechanism.}}
\label{result-table1}
\vspace{-0.15in}
\end{table*}

Since existing adversarial optimization techniques are commonly applied to achieve fair representation learning, we also investigate fairness metrics to evaluate whether our mechanism complies with various fairness standards. We consider three fairness metrics: equalized odds, equalized opportunity, and demographic parity.   In brief, \textit{equalized odds} mandates that a classifier maintains equal true positive rates and false positive rates across different sensitive groups. \textit{Equal opportunity}, a more relaxed version of \textit{equalized odds}, is satisfied if the classifier exhibits equal true positive rates for all sensitive groups, without specific consideration for false positive rates. Similarly, \textit{demographic parity}, also referred to as statistical parity, necessitates that the decision rate (the rate of predicting the positive class) remains consistent across all groups, irrespective of the base rates of the outcome within those groups \cite{garg2020fairness}. Note that, lower scores for these metrics indicate better fairness.

\subsection{Implementation Details}

\begin{figure}[!htb]
\vspace{-0.125in}
    \centering
    \includegraphics[width=0.4\textwidth]{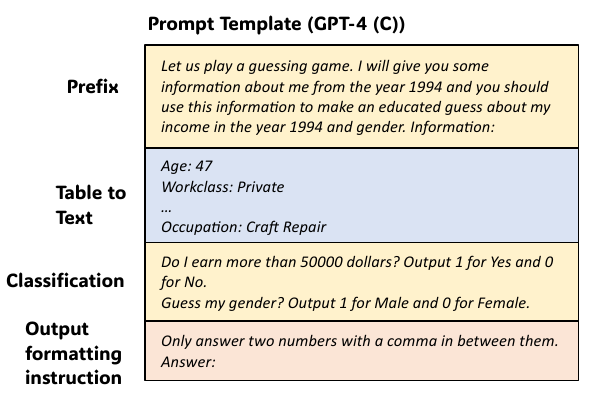}
    \caption{\textmd{GPT-4 zero-shot classification prompt, GPT-4 (C)}}
    \label{classification_prompt}
    \vspace{-0.125in}
\end{figure}

In our research, we access the GPT-4 model through OpenAI's API, specifically utilizing the \textit{gpt-4-1106-preview} model and its chat completions functionality. We utilize the default system prompt. The model is currently subject to a daily token limit of 500,000, within which our prompt equates to roughly 710 data points. Moreover, we also use GPT-4 as a zero-shot classifier. The exact prompt used for classification tasks is shown in Fig. \ref{classification_prompt}, again with the default system prompt. The inspiration for the prompt's prefix comes from \cite{staab2023memorization}, and as noted in \cite{hegselmann2023tabllm}, varying prompts don't significantly alter results, provided that all necessary information is clearly conveyed to the model. This classification prompt facilitates the use of about 2600 data points within the 500,000-token limit. Our reported results are based on a temperature parameter set to 0.1, and we found that varying this parameter did not significantly alter the core outcomes of our study.

\begin{figure*}[!t]
\centering
\minipage{0.75\textwidth}
  \includegraphics[width=0.95\textwidth, trim={0 0 2cm 0}]{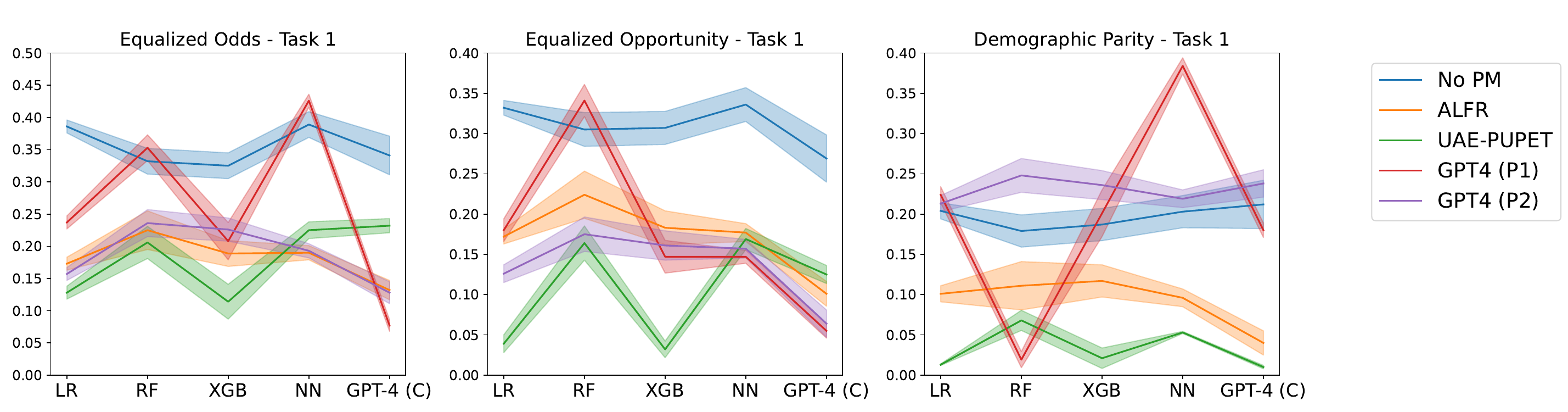}
\endminipage

\minipage{0.75\textwidth}
  \includegraphics[width=0.95\textwidth, trim={0 0 2cm 0}]{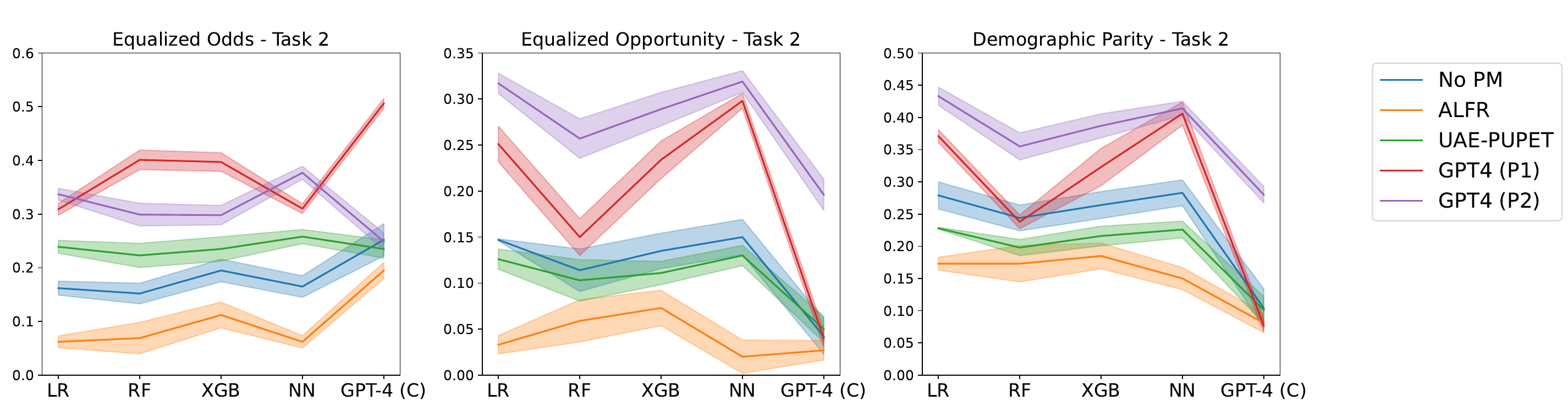}
\endminipage

\caption{\textmd{Fairness metrics for Task 1 and Task 2. The X-axis represents different ML models, and the Y-axis represents the fairness scores for those ML models, without sanitization (No PM), and after using different sanitization techniques.}}
\label{fairness}
\vspace{-0.1in}
\end{figure*}

\section{Result and Discussions}
In this section, we present the results obtained from our proposed methodology. We conduct a total of five independent experiments using different random seeds to calculate the mean and standard deviations of the results. Our test dataset comprises 1000 data points, a selection made with consideration to the associated costs with using the Open AI API and daily token limits. Table \ref{result-table1} shows the mean accuracy and F1 scores obtained by various models before and after the implementation of the sanitization mechanism for both tasks (standard deviation in the subscript). The \textit{Summary} in each segment, such as No PM, ALFR, etc., indicates the highest accuracy and F1 scores achieved among the machine learning models employed, including Logistic Regression (LR), Random Forest (RF), XGBoost (XGB), Feed Forward Neural Network (NN), and the GPT-4 Classifier (GPT-4 (C)). The highest scores are considered as the benchmark scores in inferring the private and utility features.

For Task 1, the model predicts private feature with an accuracy of 0.84 without any sanitization mechanism, and utility feature with an accuracy of 0.88. After applying ALFR and UAE-PUPET, the accuracy for private feature significantly decreases to 0.65 and 0.67 respectively, while the accuracy for utility feature remains relatively stable at 0.81 and 0.80. Our proposed method, GPT-4 (P1) - which incorporates the P1 prompt as shown in Fig. \ref{fig:gpt4-privacy overview} - yields result similar to both existing sanitization mechanisms ALFR and UAE-PUPET, reducing the accuracy for inferring private features to 0.65. For utility feature, it surpasses existing methods with an accuracy of 0.89. However, our GPT-4 (P2) approach shows a higher privacy leakage, with an accuracy of 0.75 for private feature, yet it maintains better utility (accuracy of 0.88) than other adversarial techniques. For Task 2, GPT-4 (P1) shows better privacy protection and similar utility compared to existing adversarial techniques. 
For a comparison of F1 scores and the individual results of classifier models, kindly refer Table \ref{result-table1}.

\begin{table}[!htb]
\centering
\begin{tabular}{ccccc}
\toprule
\multirow{2}{*}{\textbf{PM}} & \multicolumn{2}{c}{\textbf{Task 1}} & \multicolumn{2}{c}{\textbf{Task 2}} \\ \cmidrule(lr){2-3} \cmidrule(lr){4-5} 
                             & \textbf{$M_p$}      & \textbf{$M_u$}      & \textbf{$M_p$}      & \textbf{$M_u$}      \\ \midrule
ALFR                         & 0.00             & 0.50             & 0.07             & 0.80             \\ 
UAE-PUPET                    & 0.00             & 0.42             & 0.00             & 0.87             \\ 
GPT-4 (P1)                      & 0.00             & 1.00             & 0.00             & 0.80             \\ 
GPT-4 (P2)                      & 0.40             & 1.00             & 0.00             & 0.67             \\ \bottomrule
\end{tabular}
\caption{Privacy Leakage and Utility Performance results.}
\label{result-table2}
\vspace{-0.2in}
\end{table}

Table \ref{result-table2} showcases the performance concerning privacy leakage $(M_P)$ and utility performance $(M_U)$. For Task 1, our GPT-4 (P1) approach demonstrates result with no privacy leakage and no utility drop, surpassing the performance of existing techniques. Conversely, GPT-4 (P2) exhibits a higher degree of privacy leakage, although it maintains utility performance without a drop. For Task 2, our proposed techniques exhibit performance levels comparable to those of existing methods.

\begin{figure*}[!htb]
\centering
\minipage{0.22\textwidth}
  \includegraphics[width = \textwidth]{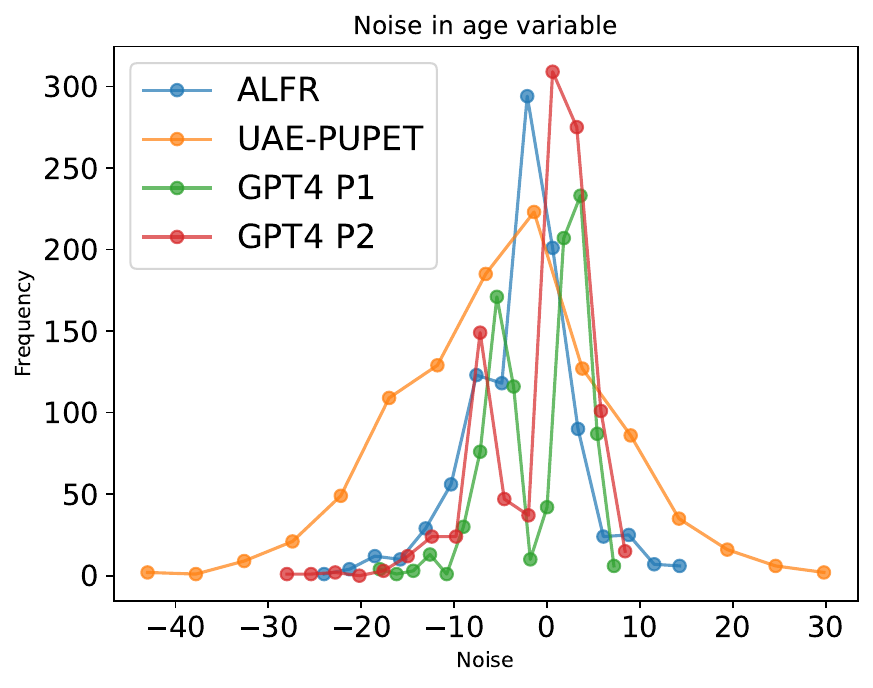}
    \label{n1}
\endminipage
\minipage{0.22\textwidth}
  \includegraphics[width = \textwidth]{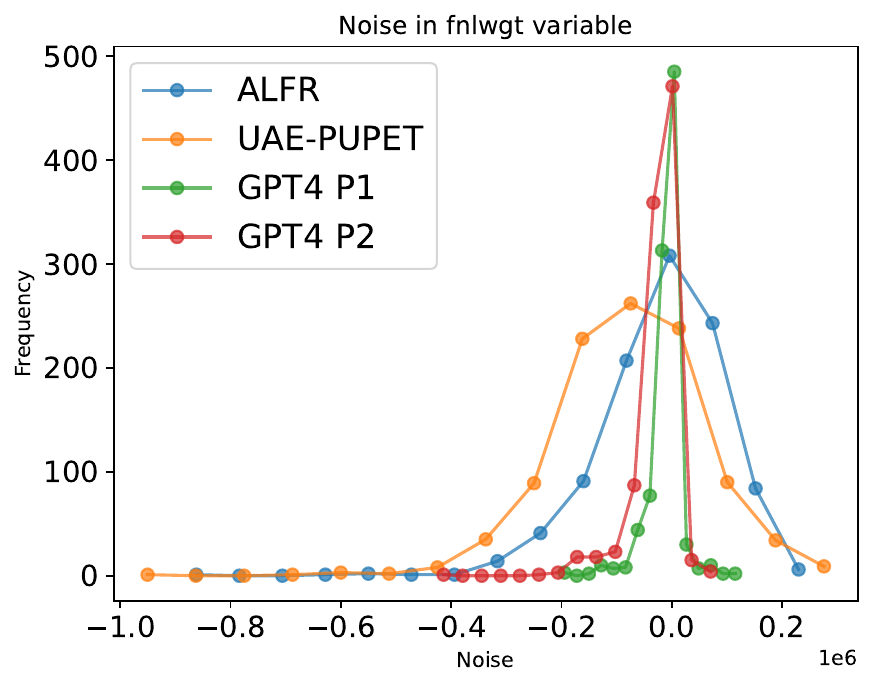}
    \label{n2}
\endminipage
\minipage{0.22\textwidth}
  \includegraphics[width = \textwidth]{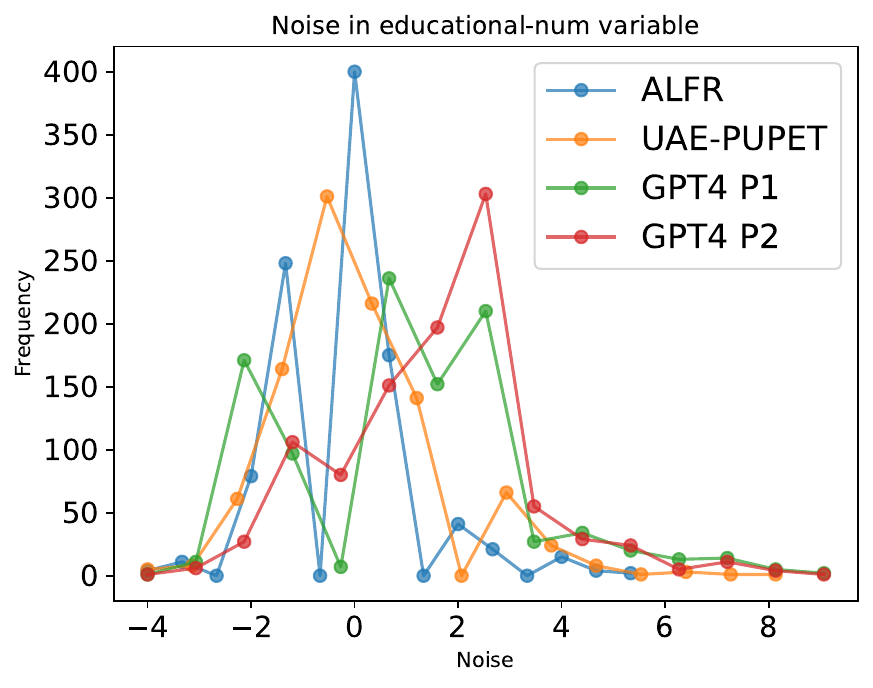}
    \label{n3}
\endminipage
\minipage{0.22\textwidth}
  \includegraphics[width = \textwidth]{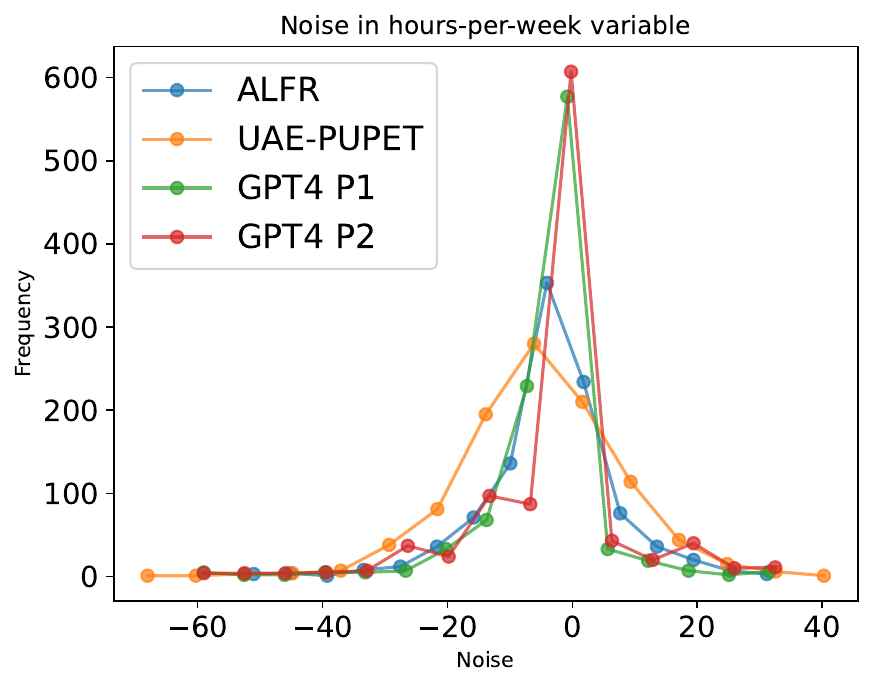}
    \label{n4}
\endminipage
\vspace{-0.15in}
\minipage{0.22\textwidth}
  \includegraphics[width = \textwidth]{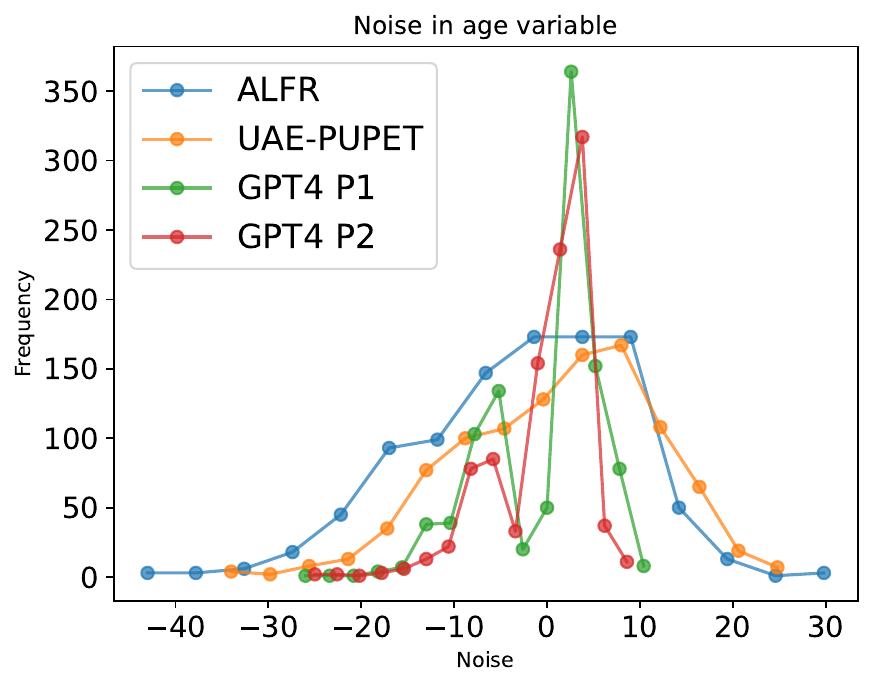}
    \label{n5}
\endminipage
\minipage{0.22\textwidth}
  \includegraphics[width = \textwidth]{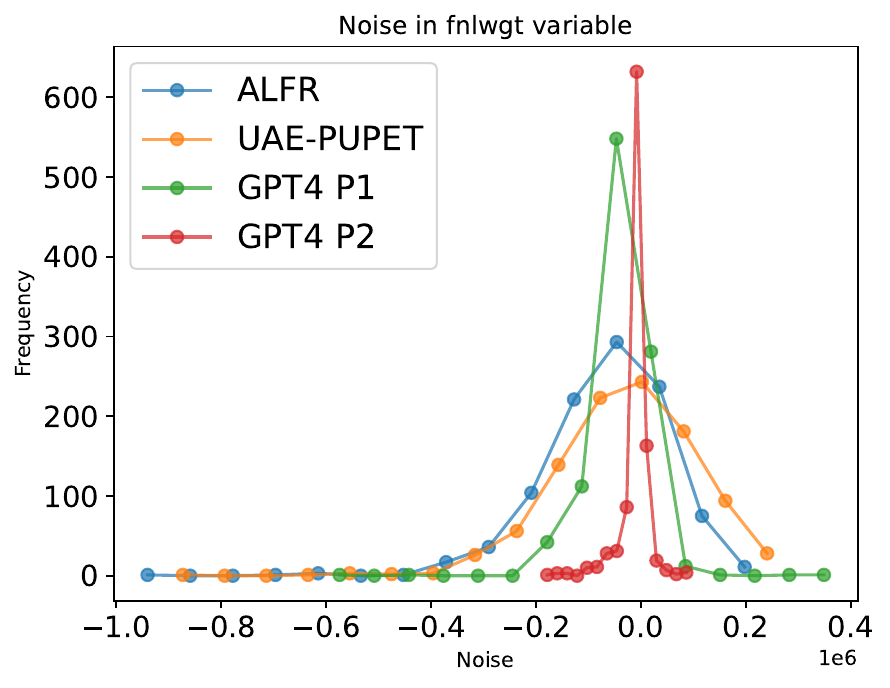}
    \label{n6}
\endminipage
\minipage{0.22\textwidth}
  \includegraphics[width = \textwidth]{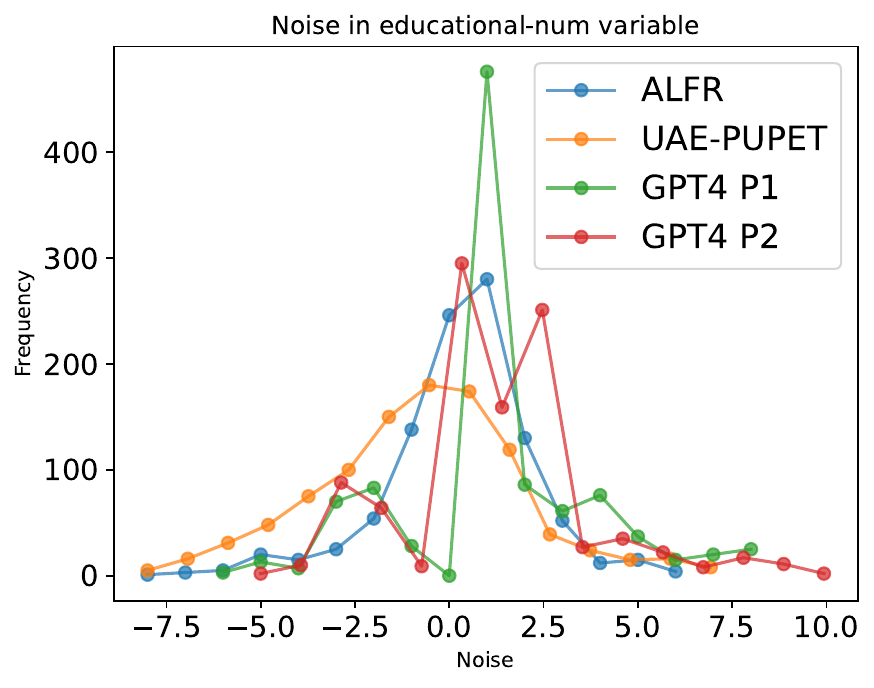}
    \label{n7}
\endminipage
\minipage{0.22\textwidth}
  \includegraphics[width = \textwidth]{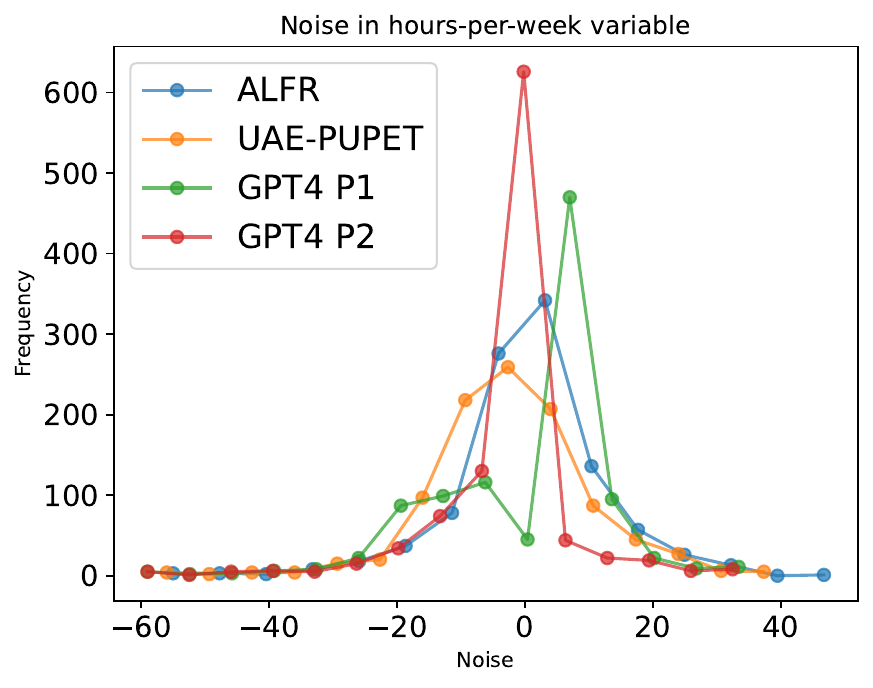}
    \label{n8}
\endminipage
\vspace{-0.15in}
\caption{\textmd{Visualization of distortion (noise) applied to continuous variables by different sanitization mechanisms for Task 1 and Task 2. Upper row shows distortion applied for Task 1 and lower row shows distortion applied for Task 2.}}
\label{fig-noise}
\vspace{-0.2in}
\end{figure*}

\begin{figure}[!htb]
\centering
\minipage{0.22\textwidth}
  \includegraphics[width = \textwidth]{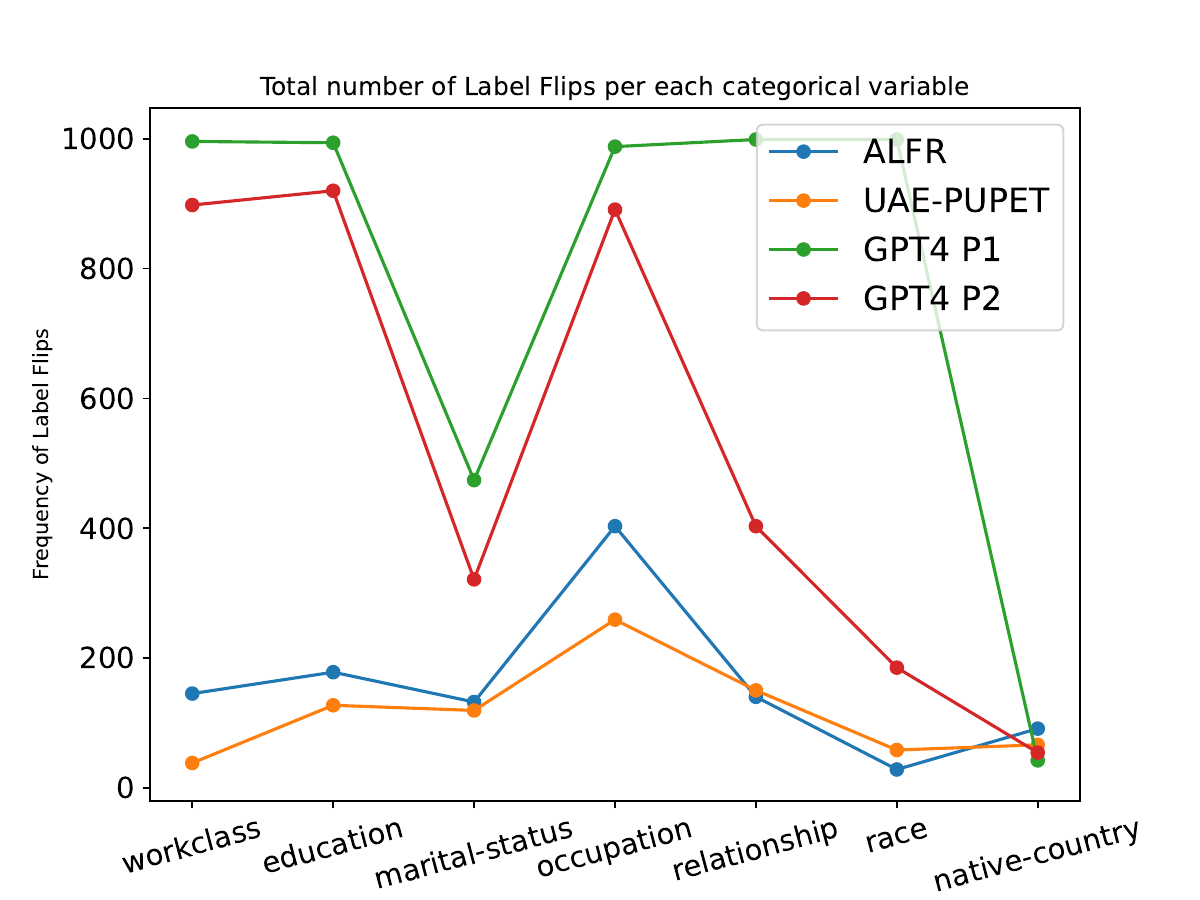}
  \subcaption{\scriptsize{Task 1}}
    \label{label-flip1}
\endminipage
\minipage{0.22\textwidth}
  \includegraphics[width = \textwidth]{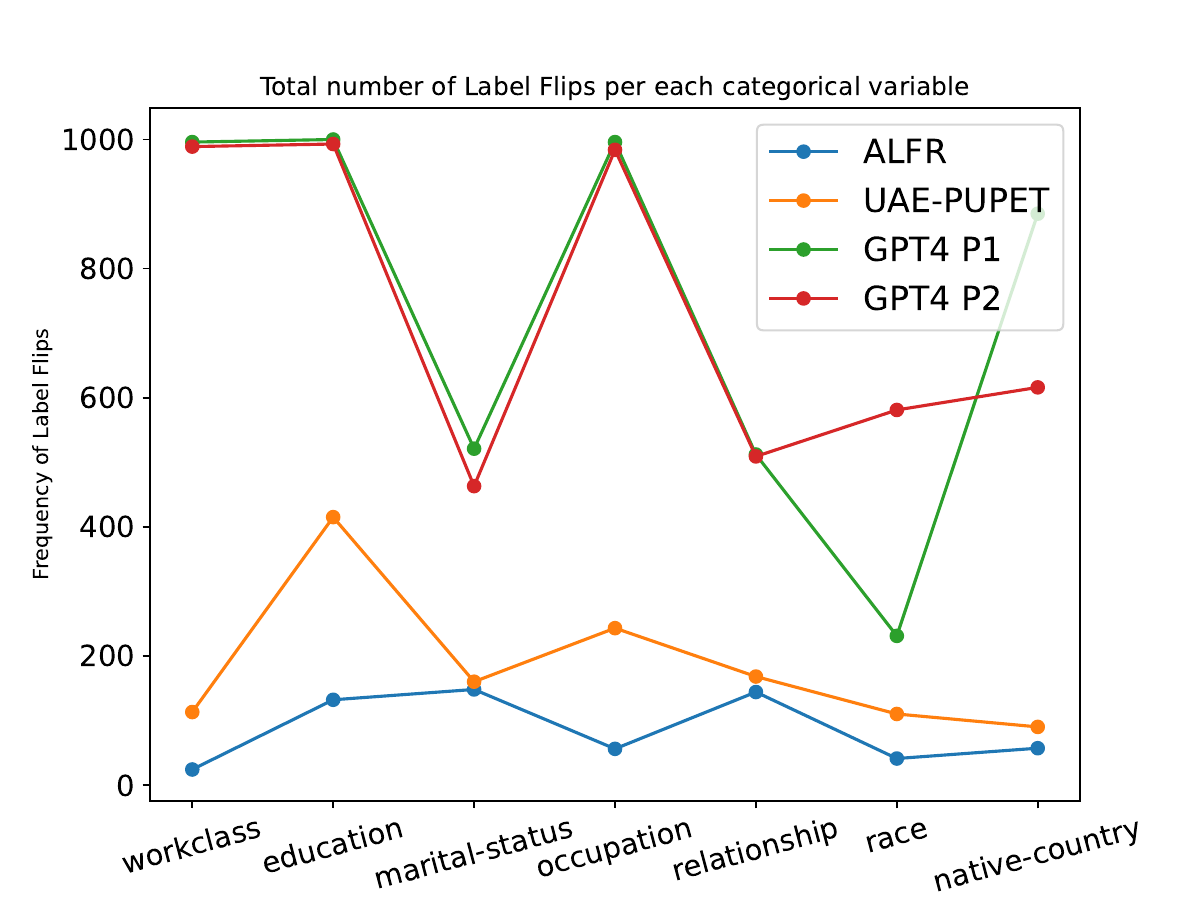}
  \subcaption{\scriptsize{Task 2}}
    \label{label-flip2}
\endminipage
\caption{\textmd{Label flips for categorical variables for both tasks.}}
\vspace{-0.15in}
\label{fig-noise2}
\end{figure}

Numerous existing optimization strategies focus on sanitizing datasets to enhance fairness in current models. It's crucial to evaluate whether our sanitization method can similarly promote fairness. Fig. \ref{fairness} depict fairness metrics for Task 1 and Task 2, respectively. Each line in these plots represent a different sanitization mechanism, with the exception of the \textit{No PM} line, which illustrates the fairness levels of various models before any sanitization. While the ALFR and UAE-PUPET methods improve fairness over not using any sanitization across all three fairness metrics, our GPT-4 (P1) mechanism falls short in providing comparable fairness. A closer examination of Prompt (P1) reveals it doesn't explicitly address fairness, which might explain its relatively lower fairness performance. However, it does show promise in certain scenarios, achieving the best fairness scores with some classification models (e.g., GPT-4 (C) for Equalized Odds and Equalized Opportunity, RF for Demographic Parity).

Since prompt P1 didn't include any mention of fairness, we introduce prompt P2, which specifically instructs the GPT-4 model to take fairness into account. This prompt shows notable improvement in Task 1 for equalized odds and equalized opportunity, achieving results on par with ALFR. However, for Task 2 (also for Task 1, demographic parity metric), the introduction of the prompt P2 does not enhance fairness, and in many cases, the scores worsen, except for the GPT-4 (C) classifier model, where fairness metrics are comparable to existing techniques.

This observation suggests that while our models don't uniformly perform similar to existing techniques in fairness metrics, there is potential in leveraging LLMs with new improved LLMs in the future.

In our study, Prompt (P1) is tailored to sanitize the dataset by reducing accuracy, while Prompt P2 is designed to improve the fairness of the models. Consistent with their objectives, our results indeed show that P1 offers enhanced privacy protection, whereas P2 better addresses fairness constraints. Additionally, we experimented with another prompt that combines elements of P1 and P2, aiming to harness the strengths of both. However, this combined prompt did not yield superior results. The specific wording of this prompt (for Task 1) was: \textit{My Income category can be correctly inferred, but my Gender cannot be correctly inferred (to fulfill fairness metrics)}. This phrase was intended to balance privacy protection and fairness enhancement, but it did not lead to the anticipated improvements in the models' performance.

Furthermore, we try to gain further understanding of the distortion (noise) introduced using our proposed methodology compared to existing techniques. Therefore, we compare the noise introduced in both existing and proposed mechanisms, to identify any potential patterns. Fig. \ref{fig-noise} illustrates the noise levels in continuous variables, excluding the variables \textit{capital-gain} and \textit{capital-loss} for the sake of brevity. The upper row of the figure depicts the noise for Task 1, while the lower row corresponds to Task 2. For continuous variables, it is observed that the noise added by our proposed mechanisms is generally less than that by the ALFR and UAE-PUPET methods, with the distributions retaining similar shapes.

In the context of categorical variables, noise is quantified as label flips -- essentially assessing whether labels change before and after the application of the sanitization mechanism. Figures \ref{label-flip1} and \ref{label-flip2} display the counts of label flips for each categorical variable in Task 1 and Task 2, respectively. We note that for categorical variables, the ALFR and UAE-PUPET methods result in fewer label flips compared to our GPT-4 (P1) and GPT-4 (P2) approaches, where for some variables almost all the categories are flipped.

Our research further seeks to determine whether actual private and utility labels are necessary for effective dataset sanitization. To explore this, we experimented with an unsupervised approach, as depicted in Fig. \ref{fig:gpt4-privacy overview}, \textit{without supervision} block, where the models were not provided with actual labels (including no range for income), to assess if they could autonomously discern the true values and sanitize the dataset accordingly. We find that including true label values, i.e., the supervised approach is important.

\begin{table}[!htb]
\centering
\begin{tabular}{ccccccc}
\toprule
                & \multicolumn{3}{c}{\textbf{Private feature}}                                     & \multicolumn{3}{c}{\textbf{Utility feature}}                                     \\ \cmidrule(lr){2-4} \cmidrule(lr){5-7}
                & \multicolumn{1}{c}{\textbf{Acc}} & \multicolumn{1}{c}{\textbf{Precision}} & \textbf{F1} & \multicolumn{1}{c}{\textbf{Acc}} & \multicolumn{1}{c}{\textbf{Precision}} & \textbf{F1} \\ \midrule
\textit{Task 1} & \multicolumn{1}{c}{0.68}         & \multicolumn{1}{c}{0.73}       & 0.69        & \multicolumn{1}{c}{0.78}         & \multicolumn{1}{c}{0.78}       & 0.78        \\ 
\textit{Task 2} & \multicolumn{1}{c}{0.69}         & \multicolumn{1}{c}{0.59}       & 0.62        & \multicolumn{1}{c}{0.60}         & \multicolumn{1}{c}{0.61}       & 0.60        \\ \bottomrule
\end{tabular}
\caption{GPT-4 (P1) sanitization without supervision.}

\label{result-table3}
\end{table}

Table \ref{result-table3} displays the highest accuracy, precision, and F1 scores achieved across the five classification models used in our study. While the outcomes for Task 1 were somewhat acceptable, the results for Task 2 clearly indicate the necessity of true labels for effective sanitization. It's also noteworthy that the highest accuracy for the utility feature in Task 1, as mentioned in the table, was achieved by only a single model. For the remaining models, there was a notable absence of preserved information for the utility feature, underscoring the challenges faced by unsupervised method in maintaining utility while protecting privacy.

\section{Conclusion}
In our study, we present an initial study of GPT-4 based sanitization mechanism that operates without the need for additional training, relying solely on the implementation of specifically designed prompts. Our findings reveal that this approach offers a level of privacy protection against various machine learning models that is comparable to existing techniques, despite its relative simplicity. While our method demonstrates efficacy in privacy protection, it does not consistently meet fairness constraints to the same extent as existing methods. However, certain results within our study do show promise, exhibiting comparable outcomes to those of established techniques. Given the rapid advancement in model development, we are optimistic that future iterations of such models will further enhance the capabilities of our proposed method, potentially addressing its current limitations in ensuring fairness.

\bibliographystyle{IEEEtran}
\bibliography{IEEEabrv, references}

\end{document}